\def\cvprPaperID{11770}
\def\confName{CVPR}
\def\confYear{2023}

\def\paperTitle{Task Bias in Vision-Language Models}

\def\authorBlock{
    Sachit Menon\thanks{Equal contribution} \quad
    Ishaan Preetam Chandratreya\footnotemark[1] \quad
    Carl Vondrick \\
    Department of Computer Science, Columbia University \\
    \texttt{\{sm4934,ipc2107,cv2428\}@columbia.edu}
}

\newif\ifreview 
\newif\ifarxiv \newcommand{\arxiv}{\arxivtrue}
\newif\ifcamera 
\newif\ifrebuttal 

\arxiv % \review OR \arxiv OR \cameraready

\pdfoutput=1
\documentclass[10pt,twocolumn,letterpaper]{article}
\ifreview \usepackage[review]{cvpr} \fi
\ifarxiv \usepackage[pagenumbers]{cvpr} \fi
\ifrebuttal \usepackage[rebuttal]{cvpr} \fi
\ifcamera \usepackage{cvpr} \fi

\usepackage{graphicx}
\usepackage{amsmath}
\usepackage{amssymb}
\usepackage{booktabs}

\usepackage{times}
\usepackage{microtype}
\usepackage{epsfig}
\usepackage[table,xcdraw]{xcolor}
\usepackage{caption}
\usepackage{float}
\usepackage{placeins}
\usepackage{color, colortbl}
\usepackage{stfloats}
\usepackage{enumitem}
\usepackage{tabularx}
\usepackage{xstring}
\usepackage{multirow}
\usepackage{xspace}
\usepackage{url}
\usepackage{xcolor}
\usepackage[hang,flushmargin]{footmisc}
\usepackage{sidecap}

\ifcamera \usepackage[accsupp]{axessibility} \fi

% ...

\ifarxiv  \fi

\newcommand{\R}[1]{{%
    \textbf{%
        \ifstrequal{#1}{1}{\textcolor{red}{R#1}}{%
        \ifstrequal{#1}{2}{\textcolor{blue}{R#1}}{%
        \ifstrequal{#1}{3}{\textcolor{magenta}{R#1}}{%
        \ifstrequal{#1}{4}{\textcolor{teal}{R#1}}{%
                           \textcolor{cyan}{R#1}%
        }}}}%
    }%
}}

\definecolor{purpobj}{HTML}{a69ab9}
\definecolor{greentxt}{HTML}{72ac6f}
\definecolor{orangeact}{HTML}{e5ac7c}

\usepackage{xr-hyper}

\makeatletter
\newcommand*{\addFileDependency}[1]{
  \typeout{(#1)}
  \@addtofilelist{#1}
  \IfFileExists{#1}{}{\typeout{No file #1.}}
}

\makeatother

\usepackage[pagebackref,breaklinks,colorlinks]{hyperref}
\usepackage[capitalize]{cleveref}
\crefname{section}{Sec.}{Secs.}
\crefname{table}{Table}{Tables}
\crefname{figure}{Fig.}{Figs.}

\usepackage{wrapfig}
\usepackage{subcaption}
\usepackage{booktabs}
\usepackage{multirow}
\begin{document}
%% TITLE
\title{\paperTitle}
\author{\authorBlock}
\maketitle

\begin{abstract}

Incidental supervision from language has become a popular approach for learning generic visual representations that can be prompted to perform many recognition tasks in computer vision. We conduct an in-depth exploration of the CLIP model and show that its visual representation is often strongly biased towards solving some tasks more than others. Moreover, which task the representation will be biased towards is unpredictable, with little consistency across images. To resolve this task bias, we show how to learn a visual prompt that guides the representation towards features relevant to their task of interest. Our results show that these visual prompts can be independent of the input image and still effectively provide a conditioning mechanism to steer visual representations towards the desired task.

\end{abstract}
\section{Introduction}

Learning visual representations from the images and text on the Internet has become a popular technique for creating visual models that solve many recognition tasks  at once \cite{radford_learning_2021,ramesh_hierarchical_2022,dhruv_paper,huypaper}. Since vision-language models (VLMs) learn the association between a text phrase and an image, a zero-shot classification paradigm has become effective where image recognition can be performed by defining the categories with free-form text.
Language has been a strong source of incidental supervision for visual representation learning.

\begin{figure}[t!]
    \centering
    \includegraphics[width=0.4\textwidth]{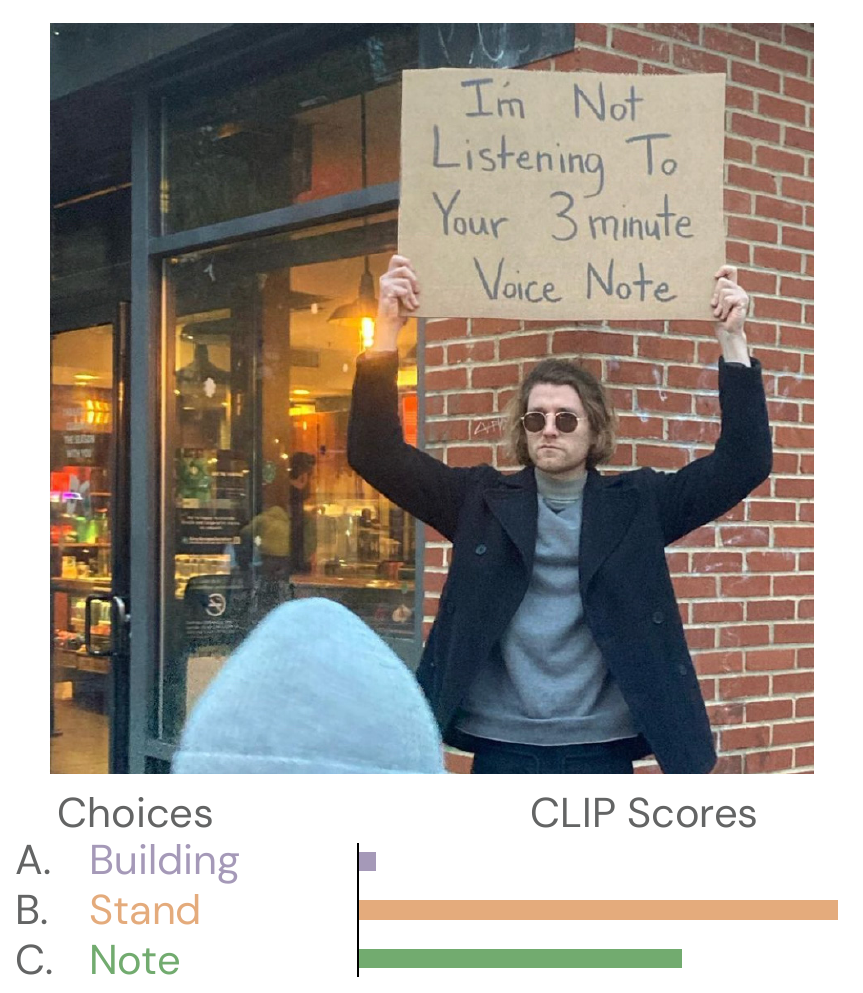}
    \caption{This paper shows that visual representations have a \textit{task bias} where they are predisposed to solving some tasks over others, and this bias even changes between images. For the above image, the CLIP visual representation is strongly biased towards solving action recognition over scene text and object recognition, even though all of the answers would be correct.}
    \label{fig:taskambimg}
\end{figure}

However, language also has risks. Fig.~\ref{fig:taskambimg} shows a typical setup for VLMs, where we have an image and a set of category labels. If you look carefully, you will see that all of these choices correspond to different tasks. Most of the image is the object \textit{building}, the action being performed is \textit{stand}, and the word \textit{note} appears as text. Which task will the VLM decide to pick? This question provides a probe into the inner workings of the VLM and which features it chooses to base its representation.

We have found that in the case of the most widely-used VLM, CLIP \cite{radford_learning_2021}, the visual representations have already decided which task to answer solely based on the image.
%; the information they contain primarily pertains to one task.
As shown on the right of Fig.~\ref{fig:taskambimg}, when CLIP is given these choices, it has a bias to prefer \textit{stand}, even if all options are equally correct.
For most images in our dataset, our results show that visual representations have a \emph{task bias} where the representation strongly prioritizes one task over others. Moreover, the bias is able to shift depending on the image in unpredictable ways. For some images, the representation is predisposed for object recognition, some for action recognition, and others for scene text recognition. 

%We identify this limitation of this zero-shot application paradigm\sm{remove this next bit?}.  We reveal that CLIP, the most widely used VLM, produces image representations strongly biased towards a particular task for a given image, often in unintuitive ways. 

\begin{figure*}
    \centering
    \includegraphics[width=0.95\linewidth]{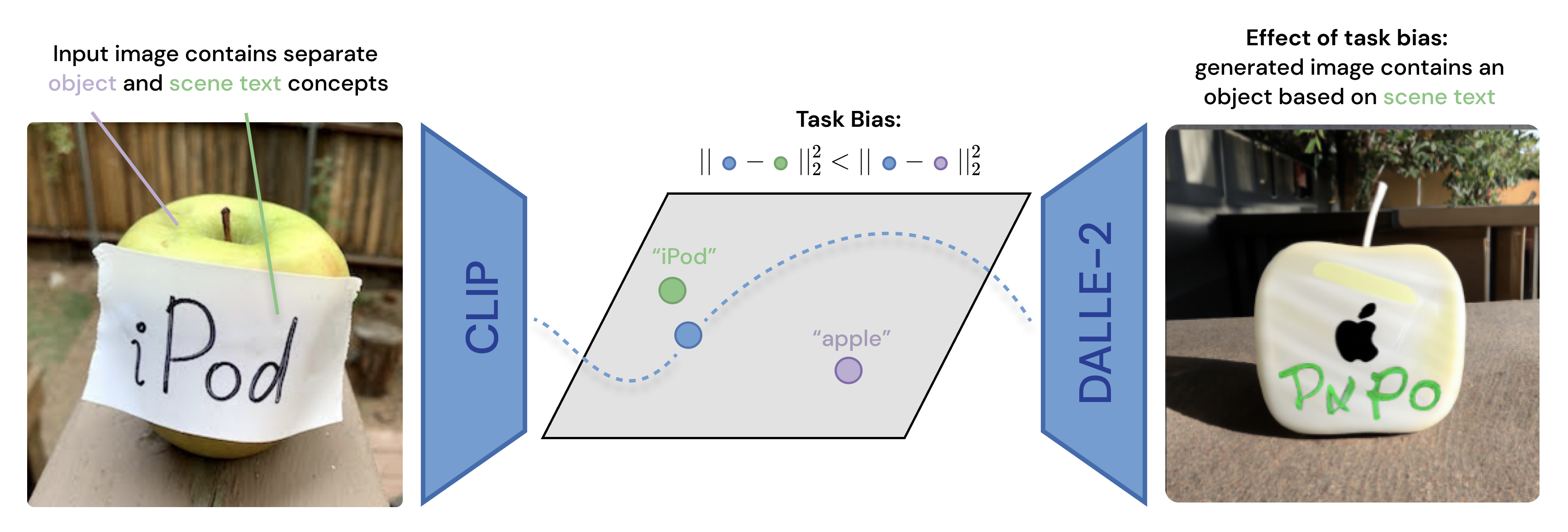}
    \caption{DALLE-2 is can act as an image generator conditioned on CLIP's visual representations. As such, it has been used as a CLIP inverter, alternatively known from unCLIP. This example, selected from \cite{ramesh_hierarchical_2022}, highlights that DALLE-2 inherits CLIP's task bias issue as the generation process depends on a CLIP representation. Despite the input image containing many different concepts (note the colors: \textcolor{greentxt}{scene text}, \textcolor{purpobj}{objects}), the decoded image principally contains objects based only on the concept propagated through \textcolor{greentxt}{scene text}. This is a consequence of task bias.}
    \vspace{-1em}
    \label{fig:dalle2}
\end{figure*}

Although the field desires visual representation that are generic, task bias highlights that these representations are more task-centric than we might imagine a priori, posing an issue for downstream use of these representations. For instance, the popular DALL-E 2 text-to-image model \cite{ramesh_hierarchical_2022} is trained to invert the CLIP image encoder, but because the image encoder can selectively encode information about certain tasks and not others, this can lead DALL-E 2 to only consider information relevant to that task in its associated generations (see Fig.~\ref{fig:dalle2}). Attempting to engineer better text prompts to clarify the task cannot fix the problem because our results show that the bias exists in the visual representation before the categorical labels are considered.

\begin{figure*}[t!]
    \centering
    \includegraphics[width=0.85\textwidth]{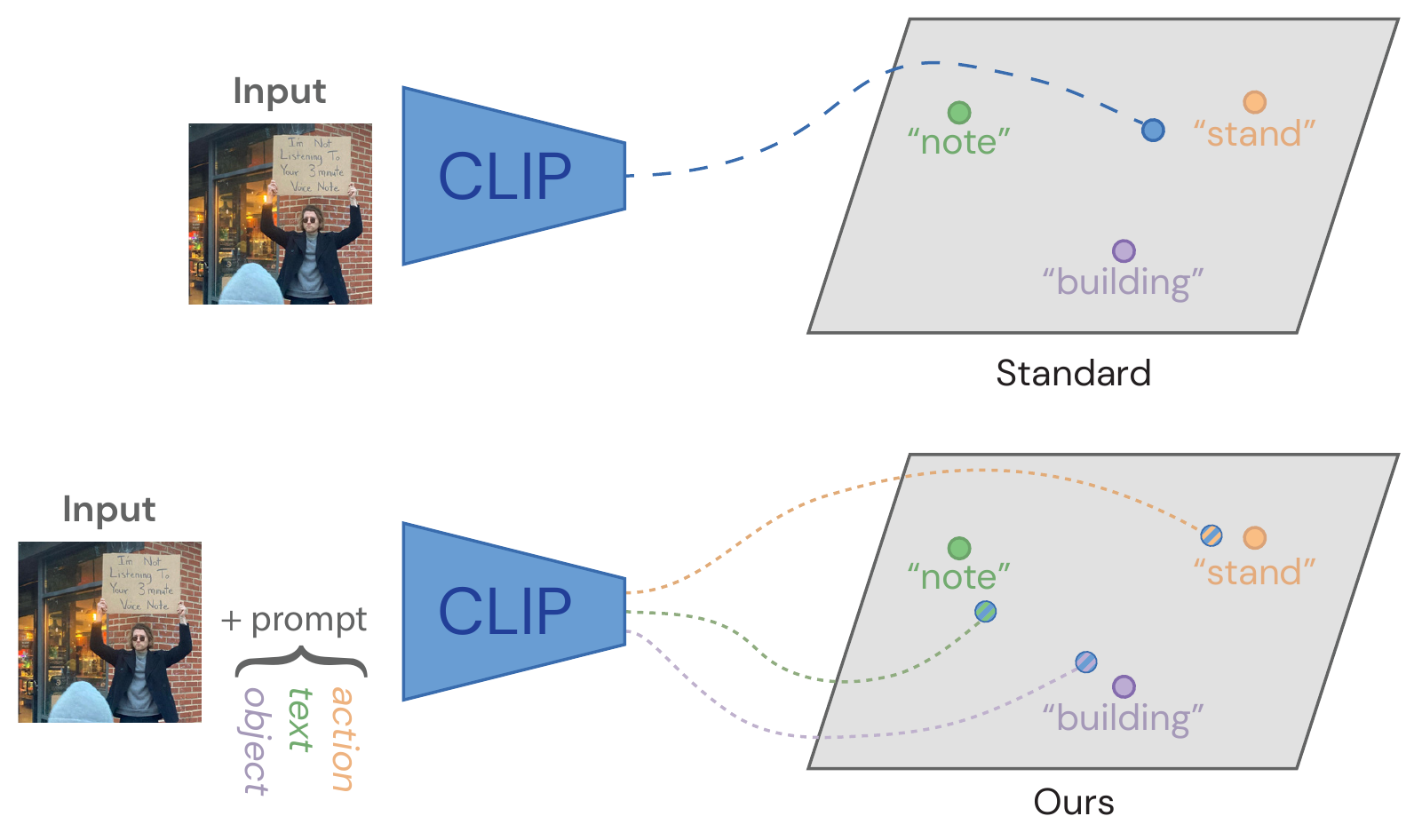}
    \caption{(top) We show standard CLIP usage where an image is embedded into a multi-modal space, and the nearest word embedding is retrieved as the final answer. However, when there are multiple correct options, this approach has a \textit{task bias} where it prefers one task over the others. (bottom) We show to overcome this problem by learning a visual prompt that steers the representation towards the intended task.}
    \label{fig:latents}
\vspace{-1em}
\end{figure*}

We propose a straightforward solution to the task bias problem that requires no modification to the model parameters, making the fix widely applicable to large-scale models. We show that we can learn a visual prompt for each task that guides the image representations towards the task of interest, thereby addressing task bias. Fig.~\ref{fig:latents} illustrates the method. The prompt is independent of the image, and experiments show it steers the visual representation to focus on the task-relevant parts of the image.

\section{Preliminaries}

For the purposes of this paper, when we discuss VLMs, we refer to the paradigm of 1) training deep neural networks using contrastive learning with image-text pairs collected from the Internet and 2) applying them in zero-shot settings to various tasks. We center our investigation around the CLIP model \cite{radford_learning_2021}. We review both of these steps below.
% \vspace{-1em}
\subsection{Vision-Language Contrastive Pretraining}

The goal of vision-language pretraining is to utilize pairs of images and text to learn effective visual representations. Recent progress in this area has been driven by advances in contrastive learning, in large part due to its computational efficacy allowing for easier scalability than exact prediction tasks \cite{radford_learning_2021}. For each batch consisting of $N$ image-text pairs, the images and text are passed through independent vision and language encoders to obtain $N$ embeddings of each modality. The contrastive pretraining task is to identify which images go with which text. This can be thought of as a classification task: for each image in the batch, classify which text option in the batch it belongs to by maximizing the cosine similarity of embeddings of corresponding pairs and minimizing it for every other combination. 

% \vspace{-1em}
\subsection{Zero-Shot Application}\label{sec:zeroshot}

We review the typical zero-shot application pipeline for CLIP. (Note that we describe the steps for selecting a text option matching a given single image; this is symmetric with selecting an image option matching a given piece of text.) First, an image is embedded using the vision encoder. Similarly, the text options for the answer to the task are embedded using the language encoder. The cosine similarity in latent space between the image embedding and all text embeddings is used as the similarity metric. The softmax function is applied to these similarities, like the logit in typical classification, and finally the text option with the highest score is returned as the output. The steps of this pipeline are often taken for granted, but as we will see, they can have unintended consequences.  
% \vspace{-1em}

\section{The Problem of Task Bias}

Training with language enables a zero-shot application procedure that can solve multiple tasks with no change in the method. However, this is a double-edged sword -- it means our visual representation can be close to a text embedding corresponding to the solution to a task we did not intend to solve. %The zero-shot application procedure only allows it to pick a single answer, regardless of how many may be ``correct'' in some sense. 
To make matters worse, we also do not know \textit{which} of the (potentially many) tasks it might be biased towards for a given decision: there is no mechanism to inform the user which task's features it is even trying to capture. %\textcolor{blue}{As an example, picking each potential 'correct' answer in Fig \ref{fig:taskbiasimg} corresponds to solving object detection, scene text recognition, activity/action recognition and scene recognition respectively. It is worth noting that many further unrelated vision tasks can also be defined on this image, and that it is equally valid for the answer to any of these tasks to form the caption of the given image.} 
In Figure \ref{fig:taskambimg}, a  zero-shot model presented with these choices could reasonably produce a visual representation close to the solutions for \textcolor{orangeact}{action recognition}, \textcolor{greentxt}{scene text recognition}, or \textcolor{purpobj}{object recognition}, depending on what parts of the input the model pays attention to, with each being valid -- and these only represent three of the potentially unbounded set of reasonable tasks that could be defined for a given image. 

\begin{figure}
    % \vspace{-2.5em}
    \centering
    \includegraphics[width=0.8\linewidth]{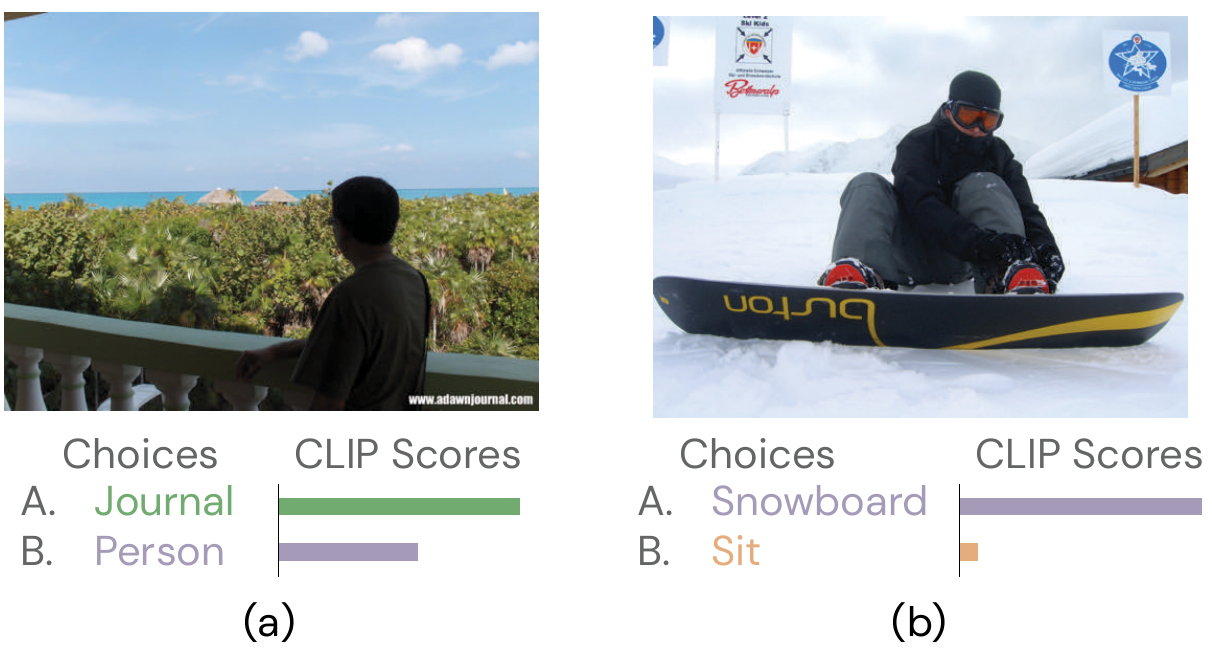}
    \caption{(a) \textcolor{greentxt}{Scene text recognition} features causing the wrong answer for \textcolor{purpobj}{object recognition}. (b) \textcolor{purpobj}{Object recognition} features causing the wrong answer for \textcolor{orangeact}{action recognition}.}
    \label{fig:snowboard}
    \vspace{-1em}
\end{figure}

We find that for given images, the visual representations CLIP produces strongly prefer one task's solution over other correct answers pertaining to the image, as discussed in Section \ref{sec:eval}. We call this tendency \textit{task bias}. Can we characterize when these preferences come into play, and to what extent? We note that while this is a form of data bias -- which is well-studied -- it is a distinct form unique to zero-shot application of VLMs that has thus far not been identified or deeply explored in its own right. 

%(This presents a flaw in our evaluation protocols for VLMs; however, the way they are practically applied often matches this zero-shot evaluation setting.) Though it is a `bias', if we can find trends in how it presents itself, we can at least make an educated guess as to what it is solving through careful experimentation. If such trends don't exist, the situation is almost worse: the task being solved would be a coin flip, inscrutable to human observers. 

\begin{figure*}[t]
    \centering
    \includegraphics[width=0.9\linewidth]{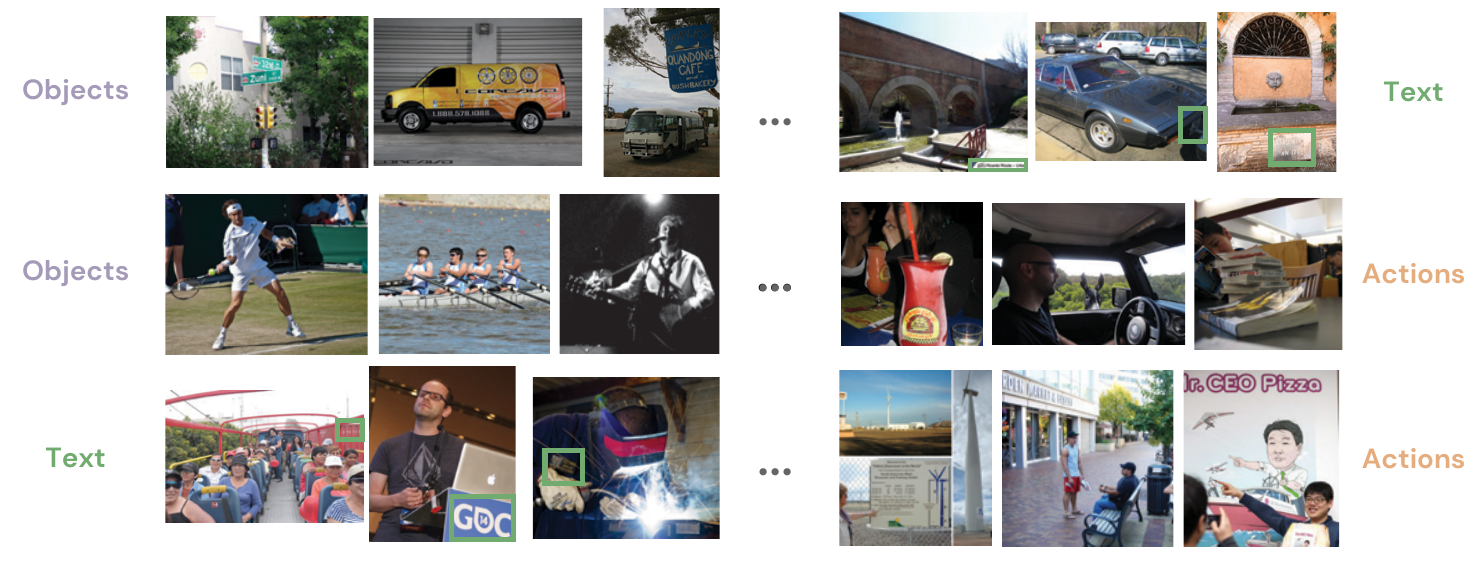}
    \caption{Example images of the most extreme per-image task bias. Best viewed with zoom. The bias is often unintuitive, picking minor text for large images (highlighted by the green boxes), obscure or small objects for clear actions, and more.\vspace{-1.5em}}
    \label{fig:bias_examples}
\end{figure*}

% On the surface, this may seem to be a frivolous concern. Knowing this, you might ask, why would you set up your choices in a way that multiple of them could be correct? This situation is not as  

For the specific application of applying CLIP-like models to a particular recognition task, this may seem on the surface to have a simple solution: try and make sure you never provide the answers to multiple questions among your options. There are two issues with this approach. First, it does not address the underlying problem with the visual representations; if the model is not paying attention to parts of the input relevant for tasks other than the one it is biased towards, the representation simply may not contain the information needed for the task. Secondly, task bias often occurs in realistic settings and is difficult to even attempt to manually address at scale.  Consider Figure \ref{fig:snowboard}. The options provided would be reasonable for the desired tasks across a dataset, but cause problems in the given images. Such examples are common when applying these models, thus making this question inevitable. In fact, this even occurs in the categories used for the ImageNet benchmark. Figure \ref{fig:swing} shows a case where CLIP picks the ImageNet category ``swing'' instead of ``baseball player,'' solving a task other than the intended one. Adding text prefixes like ``this is text that reads" does not help, which we explore further in Section \ref{sec:textprompts}.

% Our response to this question is twofold. \sm{First of all, knowing is half the battle - one of the main goals of this work is to investigate this problem precisely so that more users of these models understand it and keep it in mind while picking their options. (run-on-y sentence)} Yet the `naïve programmer' is not the only way that task bias can show up as an issue. In many cases, it is impossible to restrict your choices such that you'll never have multiple correct options in any data point, especially at scale. Consider Figure \ref{fig:} \sm{give an example where it's unavoidable to have the correct option for a different task as another choice, preferably not text.} 

This bias results in undesired behavior with these models. For example, it gives rise to the ``typographic attacks'' identified by \cite{goh_multimodal_2021}, where a model can be tricked into choosing an incorrect category for classification by instead being offered the choice of solving the \textcolor{greentxt}{scene text recognition} task as an adversary adds text to an image. We expect a new wave of work on adversarial attacks that aim to influence a VLM's choices by affecting what task's features it will select or exploiting its confusion of linguistic concepts. These can be thought of as \textit{ambiguity attacks} that exploit the ambiguity of unconditioned zero-shot recognition by influencing task bias, with ``typographic attacks'' representing the specific (interesting) case of an ambiguity attack by incentivizing the \textcolor{greentxt}{scene text recognition} task over the one the user wants.

% \vspace{-1em}
\subsection{Evaluation}
\label{sec:eval}

As task bias is a unique issue arising from zero-shot application of VLMs, we lack established ways of evaluating it. To address this, we develop a new task explicitly designed to probe task bias. Given correct text responses for multiple potential tasks, which will the visual representation from CLIP be closest to?
%(You may remember this as essentially the task you just attempted in Figure \ref{fig:taskambimg}.)
None of the answers is more true than the others; as such, there is no behavior that is ideal. \textbf{Just the fact that we can construct this task itself suggests a flaw in the way we are using VLMs.} %[\textcolor{blue}{It is worth noting that the fact that we can construct a task of this format is a criticism in itself of the existing method}]. 
Our evaluation lets us understand \textit{how} the issue presents itself rather than whether it is an issue or not. It prompts the development of new application paradigms that allow for task-conditioning in zero-shot application of VLM learned representations. %In the absence of the ability to tell a VLM what task we want it to solve, can we observe general tendencies or preferences between common tasks? 
% \textcolor{blue}{Ishaan says: we need to rephrase this as this isn't true anymore} As this is a new task, there do not yet exist datasets explicitly designed in the way that we would want \sm{somehow don't like this sentence}. Multitask datasets in vision do exist, such as the Taskonomy dataset \cite{zamir_taskonomy_2018}; however, these primarily have centered around low-level vision tasks, such as surface normal estimation, that are not as relevant for VLMs, which excel at higher-level semantic tasks. Thus, we resorted to collecting the necessary data for evaluation.

% \textcolor{blue}{For this task, we would want a dataset with a dense labelling on image properties such as that which is seen in \ref{fig:taskbiasimg}. While many existing datasets in computer vision are indeed multi-task, these tend to focus on low-level vision problems, such as surface normal estimation and texture edge detection. Instead, for the zero-shot vision-language task at hand, we would want a dataset where, for every image, there exists a set of language labels which capture distinct high-level semantic concepts represented in the image.} 

While there exist many such high level semantic tasks in natural images, we identified the following three tasks as particularly relevant, co-occurring frequently in natural images: \textcolor{purpobj}{object recognition}, \textcolor{orangeact}{action/activity recognition}, and \textcolor{greentxt}{scene text recognition}. Previous work gives us good reason to expect CLIP to perform well on each of these separately zero-shot (\cite{radford_learning_2021,goh_multimodal_2021}). We present the first work that examines CLIP's behavior when presented with all of them together.
% \begin{enumerate}
%     \item Object recognition \sm{Object/person recognition}
%     \item Action/activity recognition
%     \item Scene text recognition (OCR/reading)
%     % \item Scene recognition
% \end{enumerate}

To conduct this evaluation, we need a dataset with multiple semantic labels for each image, similar to the choices given in Figure \ref{fig:taskambimg}. Multitask datasets in vision do exist, such as the Taskonomy dataset \cite{zamir_taskonomy_2018}; however, these datasets have primarily dealt with low-level vision tasks, such as surface normal estimation, that are not as relevant for VLMs, which excel at higher-level semantic tasks.  We collected a large dataset of images where the previously mentioned tasks have clear, reliable human labels for every image. We set up our datasets to evaluate pairwise task comparisons. We obtain this data by building on the large, publicly available OpenImages-V6 repository of $\sim$ 9 million images \cite{kuznetsova_open_2020,benenson_large-scale_2019}. We obtain labels for objects by processing the provided detection labels; for actions by processing the provided labels for inter-object relationships and attributes; and for scene text by merging the images with open-source scene text labels created for a previous iteration of OpenImages \cite{krylov_open_2021}. When considering \textcolor{orangeact}{action recognition}, there are typically two ``objects'' of interest in each image -- actual objects as well as the people acting on those objects. Thus, we make our analysis of comparisons involving \textcolor{orangeact}{action recognition} more fine-grained by considering people separately from objects at large. Figure \ref{fig:snowboard} shows several example images. Please see Appendix A for further details on dataset construction.

\subsection{Establishing per-image task bias}
We probe every image by first embedding the text corresponding to correct answers for both tasks with labels, then determining which text embedding the image embedding is closer to and by how much. This follows the standard zero-shot classification protocol for VLMs (described in Section \ref{sec:zeroshot}) to do binary classification between the correct answers for both the tasks. CLIP thus assigns a score for each text option, as visualized in Figure \ref{fig:snowboard}. %For instance, suppose we were considering the image in Figure \ref{fig:taskambimg} for object recognition vs action recognition; we would present the options `weights' (the object) and `lift' (the action), obtaining a pr 
Our analysis follows from these outputs. 
%\textcolor{blue}{For every image, we follow the standard zero-shot classification protocol for VLMs to do binary classification between the correct answers for both of the tasks. This gives us (1) the answer for what CLIP chooses to do on a particular pair and (2) the probability CLIP assigned to each individual answer on a single image. } \textcolor{blue}{Using this information, we present our analysis in two steps. First, we look at what our results reveal when we preserve the task bias probabilities for individual images in our data. Then, we consider the relative frequency with which each task had its answers picked across the images in the entire dataset.}

\begin{figure}
    % \vspace{-2.5em}
    \centering
    \includegraphics[width = 0.9\linewidth]{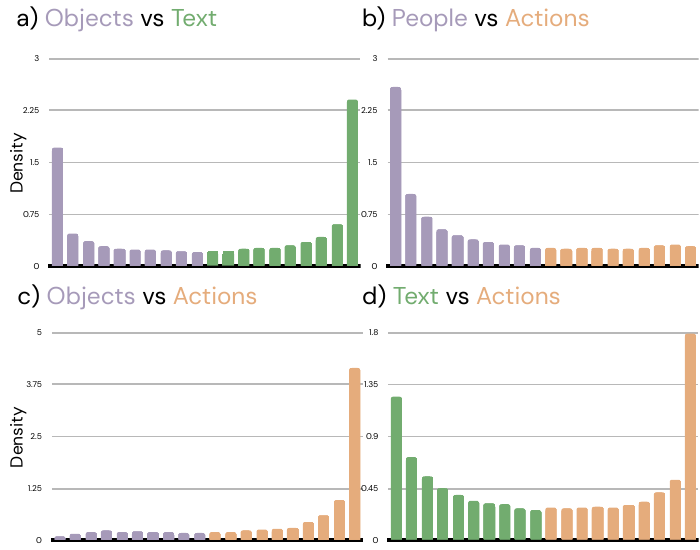}
    \caption{Task bias evaluation. We measure the distance in CLIP latent space between every image and the correct answer to both tasks. Closer to the left end implies closer to the first task listed while the right implies the second. We compute histograms of these (normalized) distances, showing that the most density can be found at both extremes. \vspace{-1.5em}}
    \label{fig:clip_prob_dist}
\end{figure}

Our results show that most embeddings are biased towards a task. Figure \ref{fig:clip_prob_dist} plots (normalized) histograms to display this trend across the datasets. We observe a distribution with highest density at the ends (max task bias). This suggests that for a given image, the predictions are strongly biased towards one particular task. One may hypothesize this bias may be predictable; for example, if an image is primarily text with a small object in it, it may be understandable to expect to solve \textcolor{greentxt}{scene text recognition}. Unfortunately, Figure \ref{fig:bias_examples} shows the opposite. We emphasize all of the answers are \textit{correct} -- but the preferred task is highly unintuitive and not aligned with what we may expect. The model ``sees'' minor objects over obvious actions, subtle actions over obvious objects, and so on.  %In particular, for two of the plots we observe a characteristic U-shape indicating that when the model embeds a particular image, it has a strong predisposition for which task it wants to solve. 
%\subsection{Is this easy to resolve?}

%How difficult is it to resolve task ambiguity? Ideally, we would be able to direct the model to solve the task we want it to solve. If we can't do this, it would at least be helpful to be able to tell which task is being solved by the model; if we wanted to do \textcolor{purpobj}{object recognition} in Figure \ref{fig:taskambimg} and the option ``lift'' were chosen, we would like to be informed that the model was doing \textcolor{orangeact}{action recognition} and not detecting an elevator. Reading the previous sections, one might think that there are some obvious approaches to resolving these issues. In this section, we demonstrate that those ``obvious'' approaches fail, emphasizing the difficulty of the problem. %\vspace{-2em}

\subsection{Text Prompt Engineering}
\label{sec:textprompts}

The first hope one might have is that we can use our existing apparatus of language prompts to indicate to the model which task we want to solve. The only immediate way the existing structure affords us to specify a task in words is through the text encoder. However, the CLIP text encoder and visual encoder are independent, only sharing an output space. Modifying our text input will only influence the text embeddings. This means our visual embedding will remain unmoved; if the visual encoder has learned to only pay attention to part of an image that solves one task (e.g., image text), the representation will only encode the information relevant to that task (e.g., text recognition), regardless of the input passed through the text encoder. 

As text prompting cannot guide our visual representations, it cannot resolve problems of task bias for downstream tasks such as image generation with CLIP embeddings (e.g. \cite{dalle2}, \cite{vqganclip}). However, we may think it could be of interest for our secondary goal: solving the intended recognition task. Say we wanted to solve \textcolor{greentxt}{scene text recognition} for Figure \ref{fig:taskambimg}. Instead of the text choices on their own, could we guide the model by using a task-specific prompt? This would add a clarifying prefix to every option -- e.g., ``This is a photo of text that reads [CHOICE]." Manual prompting has proven one of the most successful methods in guiding large language models \cite{liu_pre-train_2021,brown_language_2020}, so it seems reasonable to believe that it may help us here. To evaluate this task-directed prompting for CLIP, we created such prefixes for each task (see Appendix B for details). We added these prefixes to try and guide the decision, recomputing the respective scores. We then considered the change in proportion of data points for which the model produces the label for the intended task. We find that manual text prompts produce ambiguous results. For \textcolor{greentxt}{text} vs. \textcolor{orangeact}{actions}, the performance increases $14.7\%$ for the former, but $7.28 \%$ for the latter. For \textcolor{purpobj}{objects} vs. \textcolor{greentxt}{text}, manual text prompting gives a $9.22\%$ decrease for \textcolor{purpobj}{objects} and $11\%$ increase for \textcolor{greentxt}{text}. For \textcolor{purpobj}{objects} vs. \textcolor{orangeact}{actions}, it leads to a $37.5\%$ decrease for \textcolor{purpobj}{objects} and a $29.6\%$ increase for \textcolor{orangeact}{actions}. Overall, we do not find any consistent way to guide the model towards desired tasks using manual text prompts.

This indicates that adding manual text prompts fails to steer the task consistently and significantly. Further, it suggests that the task bias in the visual representations cannot be trivially overcome, even for the retrieval task, by simply modifying the representations of the ``answer choices.''    %We believe this motivates future work on resolving task ambiguity by integrating information from the task description with the visual embedding.

%\textcolor{blue}{Ishaan says: This paragraph seems unnecessary given that we have an entire section being prepared on context/addition of prompts} \sm{agreed. should be moved to the results section showing that this in fact doesn't work/rephrased} 

\section{Resolving Task Bias}

Given a particular task of interest, we would like to be able to change the visual representations extracted from images to primarily pertain to input features relevant to that task. We call this \textit{task disambiguation for visual representations}. We can measure this by considering, over a large set of images, how often the visual representations are closer to the solutions for the goal task than for other tasks.   %How can we influence the way that the visual encoder constructs these representations? 

In this section, we provide a simple but effective technique that allows us to reliably direct CLIP towards representations relevant to a given task without modifying the pre-trained model.  To this end, we learn a single set of parameters for each task of interest that we combine with the visual input before it is passed through the visual encoder. This can be thought of as learning a per-task visual prompt that indicates what question we are asking CLIP, guiding the information encoded in its representation.

\subsection{Method}

%CLIP's visual encoding process involves patching the input image, encoding each patch using a convolutional network into a visual token, and have a large transformer attend over a sequence of these visual tokens. Two recent works explore prompting the 
 
We are given a dataset where each input image $x$ has task labels $y_1, y_2 \ldots y_n$. For instance, in Fig. \ref{fig:taskambimg}a), we might have $y_1 = \textrm{``lift"},\quad y_2 = \textrm{``football"},\quad y_3 = \textrm{``weight"}$, where task $1$ is \textcolor{orangeact}{action recognition}, task $2$ is \textcolor{greentxt}{scene text recognition}, and task $3$ is \textcolor{purpobj}{object recognition}. Given a fixed pre-trained model parametrized by $\theta$, the goal is to learn a set of parameters $\phi$ that minimizes the objective
\begin{align}
    \max_{\phi} \sum_{y=1}^{n}  1[y=y_k] \log \frac{e^{f_{V}(g_\phi(x))\cdot f_{T} (y) }}{\sum_{k'=1}^{n} e^{f_{V}(g_\phi(x))\cdot f_{T} (y_k')}}
\end{align}
%    P_{\theta}(y_k| f_\phi(x))) \\
where $k$ is the task of interest for building our visual representations, $f_\phi$ is a function that applies the prompt parameters $\phi$ to the input image, and $f_V$ and $f_T$ are the pre-trained vision and text encoders of CLIP respectively. In other words, this is the cross-entropy of the CLIP similarity between the image and the answer to the desired task compared to every other text option. This encourages higher likelihood for the solution to task $k$ compared to those for other tasks, thereby learning a prompt that leads to a representation better suited to that task.

Several prior works explore learning prompts for visual models. However, these works deal with prompting as a method to improve the performance of pre-trained models on a particular downstream dataset, such as ImageNet \cite{bahng_exploring_2022, salman_unadversarial_2021, jia_visual_2022}. Note that in our method, unlike previous work, the label set $y_1, \ldots, y_n$ used to compute the objective differs for each instance in our batch; what remains constant is the set of tasks labeled for each associated image.

We explore two main approaches to implement the prompt application function $f_\phi$. First, we aim to learn a prompt in pixel space, similar to \cite{bahng_exploring_2022}; in this case, the learned parameters are simply added to a fixed border around the original image prior to tokenizing and image and passing it through the visual encoder. Secondly, we explore learning a prompt as a visual token, similarly to \cite{jia_pt}, where the prompt parameters are prepended to the encoded visual tokens before they are passed through the vision transformer. 

 % The method $\text{Agg}$ can be implemented using one of the many possible prior works on visual prompting. We adopt the framework of \cite{bahng_exploring_2022} for its appealing simplicity and ease of application. Here $\text{Agg}(x_i, \phi) = x_i + \phi$ where the updated parameters are simply added to a fixed number of pixels at the edges of the original image, before encoding and tokenization. We also provide results for the framework of \cite{}, where $\text{Agg}(x_i, \phi) = [\phi: \theta_{encode}(x_i)]$ where the prompt parameters are prepended to the encoded visual tokens before they are passed through the vision transformer. 

% \vspace{-1em}

\subsection{Results on Task Disambiguation}

\begin{SCtable*}[1][b!]
\centering
\begin{tabular}{p{0.1em}llrlrlr}
\toprule
& & \multicolumn{6}{c}{\textbf{Task Biases}}\\
\textbf{} & \textbf{} & \multicolumn{2}{c}{Objects vs.\ Text} & \multicolumn{2}{c}{Objects vs.\ Actions} & \multicolumn{2}{c}{Actions vs.\ Text} \\% \midrule
%\textbf{} & Direction:    & Objects            & Text            & Objects            & Actions            & Text             & Actions           \\ 
%\midrule
\cmidrule(lr){3-4}\cmidrule(lr){5-6}\cmidrule(lr){7-8}
\multirow{4}{*}{\rotatebox[origin=c]{90}{\textbf{Method}}} & No Prompt               & 50.28              & 49.72           & 48.68              & 51.32              & 43.70            & 56.30             \\
& VP (PS=1)                & 92.04              & 82.94           & 67.55              & 56.25              & 80.87            & 98.04             \\
& VP (PS=5)                & 97.67              & 95.20            & 81.13              & 96.15              & 97.59            & 98.04             \\
& ViTP                     & 99.46              & 99.15           & 100.00             & 100.00             & 100.00           & 100.00            \\ \bottomrule
\end{tabular}
\caption{\label{tab:first_numbers} Accuracy for resolving task bias on the test splits of each of our datasets using various prompting methods. Our results show that with just a few parameters (and, critically, without changing model parameters) both approaches to visual prompting successfully guide CLIP visual representations toward consistently being closer to the text solutions for one task over the other.}
\end{SCtable*}

To evaluate the effectiveness of visual prompting for resolving task bias, we construct random held-out splits from the dataset introduced above. (See Appendix A for details.) In contrast to previous sections, where we use task-specific prefixes in text label categories to allow the model to have the best chance for identifying the task associated with the word, we equalize the text prefixes for all experiments in this section. Every single option in the text retrieval set is prepended with ``This is a photo of a'' as is standard with CLIP usage. This prevents any visual prompt we learn from being optimized to simply pick the correct task-directed prompt prefix, instead of learning the task-level representation.

Our results on the held-out dataset can be found in Table \ref{tab:first_numbers}. For each task and method, we report the total percentage of data points for which the model produces the intended task label. \textbf{No Prompt} indicates the values for the unmodified zero-shot CLIP encoder. In this row, the numbers across the two tasks for each dataset must add up to $100$, as the zero-shot model will choose either of the two tasks in the dataset. In subsequent rows, which show the task-optimized results, a successful task prompted model should have higher performance for the direction in which it was prompted. Note that the numbers in all rows except the first will not add up to 100, because separate models have been trained for each task inside a dataset, and the number shown is the prompted models preference for the task it was optimized to perform. \textbf{VP} indicates visual prompting as implemented by \cite{bahng_exploring_2022}. Specifically, we use their edge prompting method, where prompt parameters replace the edges of the input image like padding, with the prompt size (PS) controlling the number of padding layers on each side. As such, the prompted model $PS = 5$ has more tuned parameters than the model with $PS = 1$. \textbf{ViTP} shows the results for vision-side prompting implemented as the typical method of prompting the inputs of transformer-style models, where parameters are added between the learned [CLS] token and input representation. We add a single embedding with the size of the model width.

We train each prompt for a \textit{single} epoch, as we find we can obtain satisfactory performance without further training. Our results indicate that by optimizing only a few parameters that are shared across all the instances in our data, it is possible to direct the model to solve our intended problem. While optimizing even a one-pixel border of parameters in pixel space on the edges of the input image can provide a significant boost in directing the model to solve the task, we find that the token-based prompting yielded the best results.

%\vspace{-1em}
\subsection{Effects on Downstream Task Performance}

We now consider how adding these task-directed prompts affects downstream performance on the associated task itself, rather than probing in comparison with other task solutions. If we have indeed led to visual representations better suited to a task, we should observe better performance on said task. We use the OpenImages object recognition categories to construct an object recognition task on our held-out dataset (see Appendix C for details); we embed all text options and evaluate with the standard zero-shot recognition procedure, choosing the category with the text embedding closest to the visual embedding in the CLIP latent space. We then compare performance, measured by classification accuracy, for this object recognition task with each of our learned prompts compared to baseline CLIP. Our results are summarized in Table \ref{tab:second_numbers}. We find that prompting towards objects improves classification performance, substantially in the case of pixel-space prompting with a border size of 1. On first glance, it is surprising that a border size of 5 achieves a lower accuracy than a border size of 1, given that the latter achieves somewhat better performance at task disambiguation. However, the increased border size has the potential to pull the input image substantially further out of distribution than the border size of 1. We hypothesize this distribution mismatch tempers the improvement derived from a more task-focused representation. 

% The results in Table ~\ref{tab:first_numbers} show that given a retrieval set containing correct answers corresponding to different tasks, we can prompt click to pick the correct answer for the right task. However, given that we are concerned about these scenarios in realistic settings, a valid question might be- what is side-effect that adding these prompts may have while picking correct answers over incorrect answers for an individual task. We test this question on the test split of our Objects v. Text dataset for the object recognition task, by adding the incorrect labels corresponding to all the categories for objects in the OpenImages dataset into the retrieval set. This is equivalent to evaluating our prompted models on a custom subset of the OpenImages dataset. Our results are summarized in Table ~\ref{tab:second_numbers}, and we show that each of our prompted methods performs better compared to the zero-shot CLIP performance on the object recognition task. 

\begin{table}
\centering
\begin{tabular}{@{}lc@{}}
\toprule
            & OpenImages Accuracy \\ \midrule
Unprompted  & 20.467          \\
VP (PS = 1) & 37.254          \\
VP (PS = 5) & 20.799          \\
ViTP        & 27.547          \\ \bottomrule
\end{tabular}
\caption{\label{tab:second_numbers} Object Recognition. } \vspace{-1em}
\end{table}
% \begin{wraptable}{rt}{0.4\linewidth}
% \begin{tabular}{@{}lc@{}}
% \toprule
%             & OpenImages Acc. \\ \midrule
% Unprompted  & 20.467          \\
% VP (PS = 1) & 37.254          \\
% VP (PS = 5) & 20.799          \\
% ViTP        & 27.547          \\ \bottomrule
% \end{tabular}
% \caption{\label{tab:second_numbers} Object Classification} \vspace{-1em}
% \end{wraptable}

\subsection{Prompting and Self-Attention}

\begin{figure}[t!]
    \centering\vspace{-1em}
    \includegraphics[width=0.9\linewidth]{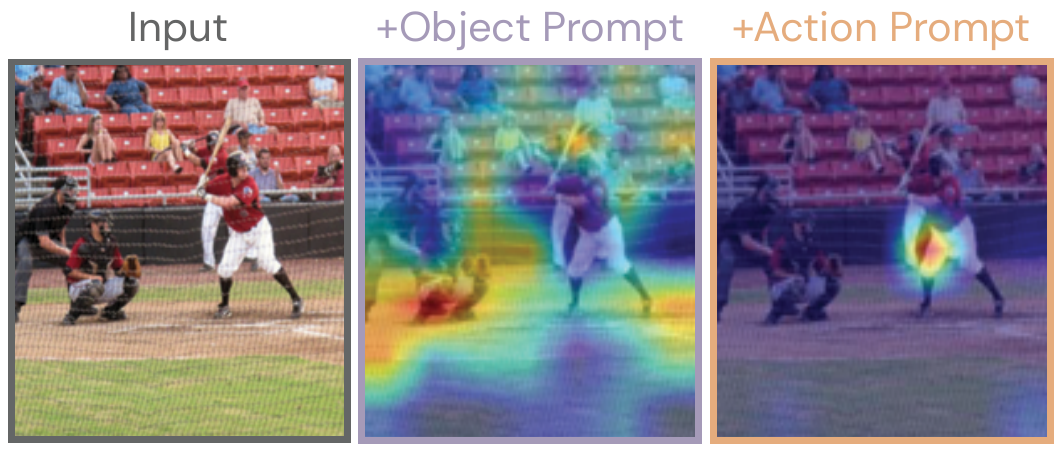}
    \caption{An example of task bias from the ImageNet dataset. CLIP chooses the ImageNet category ``swing'', as its representation of the image is biased towards actions. By applying our prompts, we can redirect the attention to the features relevant for that task.}
    \label{fig:swing}
    \vspace{-1.5em}
\end{figure}

Our goal is to modify the visual representation to direct it to capture the specific features in the image relevant to the intended task. In a transformer's encoder, this means that the token corresponding to the image embedding should pay more attention to the visual tokens of the patches most relevant for the task. Even though prompting is a light-weight, quick-to-tune modification to the input encoder, it has the ability to indirectly control the attention distribution for the rest of the image. This is because the image embedding will now also attend to the learnt prompt, changing the scale of attention to the rest of the image. We highlight the difference in attention between prompting for \textcolor{purpobj}{object recognition} and prompting for \textcolor{orangeact}{action recognition} for an example from the ImageNet dataset, which is out of distribution for our trained prompts, in Fig. \ref{fig:swing}. We see that the prompting changes the attention to put more weight on input features more relevant to the task associated with the prompt. This result highlights the fact that the prompt does not modify the attention in a fixed manner, for example to a particular part of the image or shifted by a constant offset. Rather, the prompt dynamically interacts with the content of the image to reflect semantic understanding of what kinds of features are associated with a given task, and how to find these.

\begin{figure*}[t!]
    \centering
    \includegraphics[width=\linewidth]{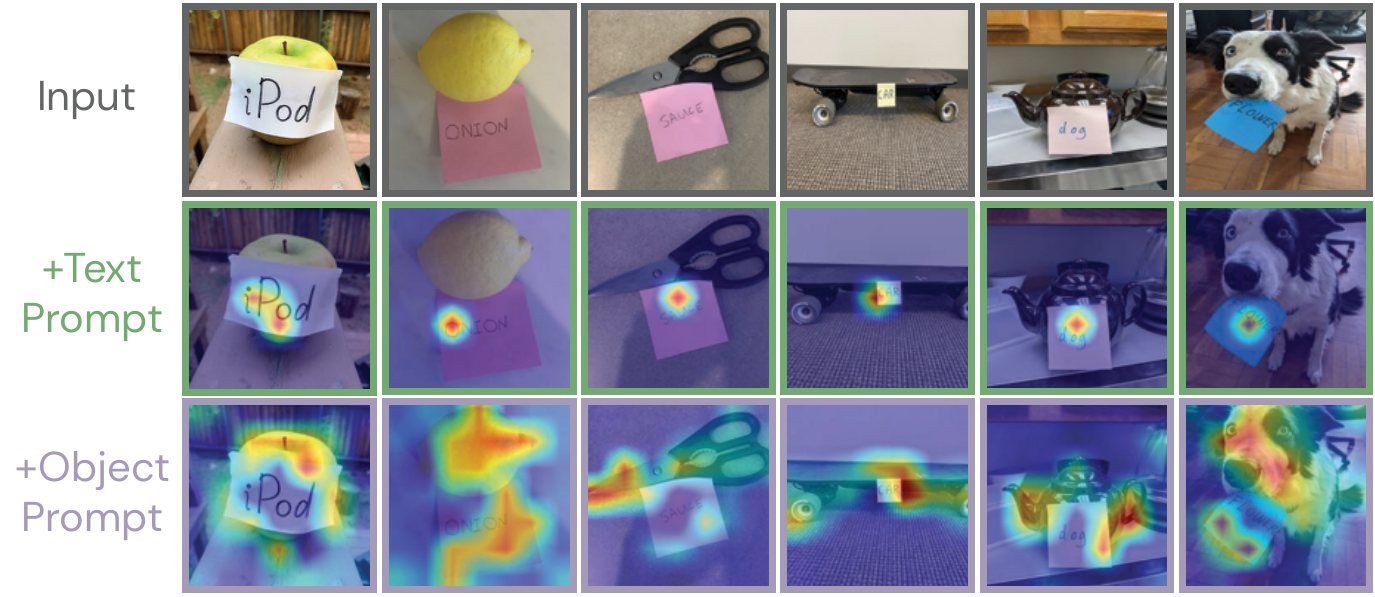}
    \caption{Difference between attention maps with \textcolor{greentxt}{scene text recognition} and \textcolor{purpobj}{object recognition} prompting on in-the-wild images from an out-of-distribution dataset \cite{samir_paper}. Best viewed with zoom. The first image is the well-known ``iPod vs Apple'' from \cite{goh_multimodal_2021}. We see the \textcolor{greentxt}{text recognition} prompt results in more attention on text and the \textcolor{purpobj}{object recognition} prompt results in more attention on surrounding objects.}
    \label{fig:attention_directed}
\end{figure*}

To test the impact of prompting on visual attention for a broader set of out-of-distribution images, we take images from \cite{goh_multimodal_2021} and the dataset collected by \cite{samir_paper}, both of which contain pictures of objects labelled with sticky notes of unrelated text. For each image, we compute two task-directed aggregated attention maps using the models that we prompt toward the respective tasks. Our method for calculating the attention maps is based on \cite{abnar_quantifying_2020,gildenblat_explainability_2021}. In order to isolate the features that are especially task relevant, we visualize the difference in the task-directed attention maps in the direction of the desired task. Our results can be seen in Figure ~\ref{fig:attention_directed}. The attention maps in the text direction clearly show a focus on the sticky notes; those extracted from the model prompted towards the objects reveal a more object-centric focus. 

% \vspace{-1.5em}

\section{Related Work}
\label{sec:related_work}

\textbf{Vision-Language Models (and Applications):} Internet pre-training with vision-and-language has become an established paradigm with the release of CLIP \cite{radford_learning_2021}, and larger models have since followed \cite{align, florence, flamingo, flava}. While designed for retrieval, the generality of the visual and text representations of CLIP has found applications in video retrieval \cite{luo_clip4clip_2021}, robotic manipulation and navigation \cite{shridhar_cliport_2021, clip_on_wheels, clip_ai}, OOD image detection \cite{mukhoti_calibrating_2020}, and image generation, to steer generative models \cite{vqganclip,gal_stylegan-nada_2021}, and much more, where parts of guiding signal for training are often based solely on the visual representation. Our work considers an inherent bias that these visual representations have toward certain tasks, and proposes a method to correct them. Visual representations free of task bias are important not only for zero shot application of CLIP in retrieval settings but also for the described downstream tasks, where a representation biased toward a particular task may lead to incorrect targets for robotic systems, misguided image generations or faulty re-identifications when used as part of safety critical applications.

\textbf{Investigating CLIP's representations:} While internet pretrained representations have seen impressive performances on several benchmarks, they are still susceptible to the most common problems in machine learning, especially deep learning in computer vision, with issues in interpretability, representation of uncertainty, social bias and robustness \cite{rudin_interpretable_2021, nado_uncertainty_2021}. \cite{goh_multimodal_2021} studies some of these issues in great detail for CLIP and introduces the notion of a \textit{typographic attack}, which shows that CLIP models are susceptible to ``reading'' instead of object detection. Subsequent works propose solutions for this specific problem: \cite{torralba_paper} consider object and text representations as normal directions in CLIP's visual representation space and learn projections on top of CLIP representations with an orthogonality constraint to isolate CLIP's reading ability. \cite{samir_paper} introduce a method to improve accuracy on finetuned tasks while preserving zero shot performance on other tasks- they explore their method in the \textit{typographic attack} setting and show that it is possible to steer CLIP away from typographic attacks with fine-tuning the entire model. Both \cite{samir_paper} and \cite{torralba_paper} use synthetic data for adapting their models, which cannot be generated outside the specific object / scene text setting. Instead, we identify typographic attacks as a specific instantiation of the broader pose a method that can alter the visual representations for specific tasks by tuning only a few parameters.  

\textbf{Learned Prompting} is a lightweight alternative to full fine-tuning used to quickly adapt pre-trained models to application domain data distributions. It has previously been used extensively in the natural language community \cite{prompttext, prompttext2} with transformers. Learned prompts are essentially a small set of parameters that are optimized for a downstream task, and can be added anywhere inside a trained model. Our work builds on \cite{bahng_exploring_2022} who show that adding prompt parameters directly to the input image is an effective alternative to linear probing and fine-tuning for CLIP, with possible improvements to robustness. Our work also builds on \cite{jia_pt} who show that adding prompts before the first self-attention layer but after encoding the inputs is also effective for vision transformers, as it is in NLP literature. Both these papers explore using vision side prompting for adapting pre-trained models to new downstream datasets; instead, we explore prompting for directing the model to perform specific tasks over others. These previous results can be considered a form of benign overfitting similar to unadversarial examples \cite{salman_unadversarial_2021}, while we produce semantically relevant, task-specific visual features. 
%\footnote{We focus on CLIP here once again, as other models -- not being open-sourced -- have yet to see the same impact.}% This potential for widespread use has even led some to dub it a ``foundation model" \cite{bommasani_opportunities_2021}, predicting that it will have substantial societal impact. In the months since CLIP was released, it has found application in video retrieval \cite{luo_clip4clip_2021}, robotic manipulation \cite{shridhar_cliport_2021}, OOD image detection \cite{mukhoti_calibrating_2020}, and more. One popular area which CLIP has been applied in is image generation, especially to steer generative models \cite{noauthor_zero-shot_nodate,noauthor_nerdyrodentvqgan-clip_nodate,gal_stylegan-nada_2021}. Outside of academic contexts, many developers are already using CLIP towards their own projects. Many projects describe bespoke search engines, image retrieval systems, or image classifiers powered by CLIP \cite{gautam_build_2021,dunteman_clip_nodate,solawetz_how_2021}. Various products and services are already being built on this technology, even including a REST API designed to for easy productionization of other CLIP-based products \cite{dunteman_clip_nodate}. This popularity underscores the need for better understanding of its strengths and flaws.

\section{Conclusions}

We explore the task bias in the representation of VLMs, finding the representation for a given image is strongly biased towards the text solution for a particular task a priori. To this end, we introduce a dataset illustrating these issues, suitable for the community to further analyze the new phenomenon. Finally, we show effective methods for guiding representations towards a task of interest via visual prompting, and show that this can substantially improve downstream recognition performance.

\textbf{Acknowledgements:} This research is based on work partially supported by the NSF NRI Award \#2132519, and the DARPA KAIROS, MCS, and SAIL-ON programs. SM is supported by the NSF Graduate Research Fellowship.

{\small
\bibliographystyle{ieee_fullname}
\bibliography{references,ishaan}

\begin{thebibliography}{10}\itemsep=-1pt

\bibitem{abnar_quantifying_2020}
Samira Abnar and Willem Zuidema.
\newblock Quantifying {Attention} {Flow} in {Transformers}.
\newblock {\em arXiv:2005.00928 [cs]}, May 2020.
\newblock arXiv: 2005.00928.

\bibitem{flamingo}
Jean-Baptiste Alayrac, Jeff Donahue, Pauline Luc, Antoine Miech, Iain Barr,
  Yana Hasson, Karel Lenc, Arthur Mensch, Katie Millican, Malcolm Reynolds,
  Roman Ring, Eliza Rutherford, Serkan Cabi, Tengda Han, Zhitao Gong, Sina
  Samangooei, Marianne Monteiro, Jacob Menick, Sebastian Borgeaud, Andrew
  Brock, Aida Nematzadeh, Sahand Sharifzadeh, Mikolaj Binkowski, Ricardo
  Barreira, Oriol Vinyals, Andrew Zisserman, and Karen Simonyan.
\newblock Flamingo: a visual language model for few-shot learning, 2022.

\bibitem{bahng_exploring_2022}
Hyojin Bahng, Ali Jahanian, Swami Sankaranarayanan, and Phillip Isola.
\newblock Exploring {Visual} {Prompts} for {Adapting} {Large}-{Scale} {Models},
  June 2022.
\newblock Number: arXiv:2203.17274 arXiv:2203.17274 [cs].

\bibitem{benenson_large-scale_2019}
Rodrigo Benenson, Stefan Popov, and Vittorio Ferrari.
\newblock Large-scale interactive object segmentation with human annotators.
\newblock {\em arXiv:1903.10830 [cs]}, Apr. 2019.
\newblock arXiv: 1903.10830.

\bibitem{opencv_library}
G. Bradski.
\newblock {The OpenCV Library}.
\newblock {\em Dr. Dobb's Journal of Software Tools}, 2000.

\bibitem{brown_language_2020}
Tom~B. Brown, Benjamin Mann, Nick Ryder, Melanie Subbiah, Jared Kaplan,
  Prafulla Dhariwal, Arvind Neelakantan, Pranav Shyam, Girish Sastry, Amanda
  Askell, Sandhini Agarwal, Ariel Herbert-Voss, Gretchen Krueger, Tom Henighan,
  Rewon Child, Aditya Ramesh, Daniel~M. Ziegler, Jeffrey Wu, Clemens Winter,
  Christopher Hesse, Mark Chen, Eric Sigler, Mateusz Litwin, Scott Gray,
  Benjamin Chess, Jack Clark, Christopher Berner, Sam McCandlish, Alec Radford,
  Ilya Sutskever, and Dario Amodei.
\newblock Language {Models} are {Few}-{Shot} {Learners}.
\newblock {\em arXiv:2005.14165 [cs]}, July 2020.
\newblock arXiv: 2005.14165.

\bibitem{vqganclip}
Katherine Crowson, Stella Biderman, Daniel Kornis, Dashiell Stander, Eric
  Hallahan, Louis Castricato, and Edward Raff.
\newblock Vqgan-clip: Open domain image generation and editing with natural
  language guidance, 2022.

\bibitem{clip_on_wheels}
Samir~Yitzhak Gadre, Mitchell Wortsman, Gabriel Ilharco, Ludwig Schmidt, and
  Shuran Song.
\newblock Clip on wheels: Zero-shot object navigation as object localization
  and exploration, 2022.

\bibitem{gal_stylegan-nada_2021}
Rinon Gal, Or Patashnik, Haggai Maron, Gal Chechik, and Daniel Cohen-Or.
\newblock {StyleGAN}-{NADA}: {CLIP}-{Guided} {Domain} {Adaptation} of {Image}
  {Generators}.
\newblock {\em arXiv:2108.00946 [cs]}, Aug. 2021.
\newblock arXiv: 2108.00946.

\bibitem{gildenblat_explainability_2021}
Jacob Gildenblat.
\newblock Explainability for {Vision} {Transformers} (in {PyTorch}), Nov. 2021.
\newblock original-date: 2020-12-29T11:27:52Z.

\bibitem{goh_multimodal_2021}
Gabriel Goh, Nick~Cammarata †, Chelsea~Voss †, Shan Carter, Michael Petrov,
  Ludwig Schubert, Alec Radford, and Chris Olah.
\newblock Multimodal {Neurons} in {Artificial} {Neural} {Networks}.
\newblock {\em Distill}, 6(3):e30, Mar. 2021.

\bibitem{huypaper}
Huy Ha and Shuran Song.
\newblock Semantic abstraction: Open-world 3d scene understanding from 2d
  vision-language models, 2022.

\bibitem{he_deep_2015}
Kaiming He, Xiangyu Zhang, Shaoqing Ren, and Jian Sun.
\newblock Deep {Residual} {Learning} for {Image} {Recognition}.
\newblock {\em arXiv:1512.03385 [cs]}, Dec. 2015.
\newblock arXiv: 1512.03385.

\bibitem{prompttext}
Yun He, Huaixiu~Steven Zheng, Yi Tay, Jai Gupta, Yu Du, Vamsi Aribandi, Zhe
  Zhao, YaGuang Li, Zhao Chen, Donald Metzler, Heng-Tze Cheng, and Ed~H. Chi.
\newblock Hyperprompt: Prompt-based task-conditioning of transformers, 2022.

\bibitem{samir_paper}
Gabriel Ilharco, Mitchell Wortsman, Samir~Yitzhak Gadre, Shuran Song, Hannaneh
  Hajishirzi, Simon Kornblith, Ali Farhadi, and Ludwig Schmidt.
\newblock Patching open-vocabulary models by interpolating weights, 2022.

\bibitem{align}
Chao Jia, Yinfei Yang, Ye Xia, Yi{-}Ting Chen, Zarana Parekh, Hieu Pham,
  Quoc~V. Le, Yun{-}Hsuan Sung, Zhen Li, and Tom Duerig.
\newblock Scaling up visual and vision-language representation learning with
  noisy text supervision.
\newblock {\em CoRR}, abs/2102.05918, 2021.

\bibitem{jia_scaling_2021}
Chao Jia, Yinfei Yang, Ye Xia, Yi-Ting Chen, Zarana Parekh, Hieu Pham, Quoc~V.
  Le, Yunhsuan Sung, Zhen Li, and Tom Duerig.
\newblock Scaling {Up} {Visual} and {Vision}-{Language} {Representation}
  {Learning} {With} {Noisy} {Text} {Supervision}.
\newblock {\em arXiv:2102.05918 [cs]}, June 2021.
\newblock arXiv: 2102.05918.

\bibitem{jia_visual_2022}
Menglin Jia, Luming Tang, Bor-Chun Chen, Claire Cardie, Serge Belongie, Bharath
  Hariharan, and Ser-Nam Lim.
\newblock Visual {Prompt} {Tuning}, July 2022.
\newblock Number: arXiv:2203.12119 arXiv:2203.12119 [cs].

\bibitem{jia_pt}
Menglin Jia, Luming Tang, Bor-Chun Chen, Claire Cardie, Serge Belongie, Bharath
  Hariharan, and Ser-Nam Lim.
\newblock Visual prompt tuning, 2022.

\bibitem{clip_ai}
Apoorv Khandelwal, Luca Weihs, Roozbeh Mottaghi, and Aniruddha Kembhavi.
\newblock Simple but effective: {CLIP} embeddings for embodied {AI}.
\newblock {\em CoRR}, abs/2111.09888, 2021.

\bibitem{kingma_adam_2017}
Diederik~P. Kingma and Jimmy Ba.
\newblock Adam: {A} {Method} for {Stochastic} {Optimization}.
\newblock {\em arXiv:1412.6980 [cs]}, Jan. 2017.
\newblock arXiv: 1412.6980.

\bibitem{krylov_open_2021}
Ilya Krylov, Sergei Nosov, and Vladislav Sovrasov.
\newblock Open {Images} {V5} {Text} {Annotation} and {Yet} {Another} {Mask}
  {Text} {Spotter}.
\newblock {\em arXiv:2106.12326 [cs]}, June 2021.
\newblock arXiv: 2106.12326.

\bibitem{kuznetsova_open_2020}
Alina Kuznetsova, Hassan Rom, Neil Alldrin, Jasper Uijlings, Ivan Krasin, Jordi
  Pont-Tuset, Shahab Kamali, Stefan Popov, Matteo Malloci, Alexander
  Kolesnikov, Tom Duerig, and Vittorio Ferrari.
\newblock The {Open} {Images} {Dataset} {V4}: {Unified} image classification,
  object detection, and visual relationship detection at scale.
\newblock {\em International Journal of Computer Vision}, 128(7):1956--1981,
  July 2020.
\newblock arXiv: 1811.00982.

\bibitem{liu_pre-train_2021}
Pengfei Liu, Weizhe Yuan, Jinlan Fu, Zhengbao Jiang, Hiroaki Hayashi, and
  Graham Neubig.
\newblock Pre-train, {Prompt}, and {Predict}: {A} {Systematic} {Survey} of
  {Prompting} {Methods} in {Natural} {Language} {Processing}.
\newblock {\em arXiv:2107.13586 [cs]}, July 2021.
\newblock arXiv: 2107.13586.

\bibitem{luo_clip4clip_2021}
Huaishao Luo, Lei Ji, Ming Zhong, Yang Chen, Wen Lei, Nan Duan, and Tianrui Li.
\newblock {CLIP4Clip}: {An} {Empirical} {Study} of {CLIP} for {End} to {End}
  {Video} {Clip} {Retrieval}.
\newblock {\em ArXiv}, 2021.

\bibitem{dhruv_paper}
Arjun Majumdar, Gunjan Aggarwal, Bhavika Devnani, Judy Hoffman, and Dhruv
  Batra.
\newblock Zson: Zero-shot object-goal navigation using multimodal goal
  embeddings, 2022.

\bibitem{torralba_paper}
Joanna Materzynska, Antonio Torralba, and David Bau.
\newblock Disentangling visual and written concepts in clip, 2022.

\bibitem{fiftyone}
B.~E. Moore and J.~J. Corso.
\newblock Fiftyone.
\newblock {\em GitHub. Note: https://github.com/voxel51/fiftyone}, 2020.

\bibitem{mukhoti_calibrating_2020}
Jishnu Mukhoti, Viveka Kulharia, Amartya Sanyal, Stuart Golodetz, Philip H.~S.
  Torr, and Puneet~K. Dokania.
\newblock Calibrating {Deep} {Neural} {Networks} using {Focal} {Loss}.
\newblock {\em arXiv:2002.09437 [cs, stat]}, Oct. 2020.
\newblock arXiv: 2002.09437.

\bibitem{nado_uncertainty_2021}
Zachary Nado, Neil Band, Mark Collier, Josip Djolonga, Michael~W. Dusenberry,
  Sebastian Farquhar, Angelos Filos, Marton Havasi, Rodolphe Jenatton, Ghassen
  Jerfel, Jeremiah Liu, Zelda Mariet, Jeremy Nixon, Shreyas Padhy, Jie Ren, Tim
  G.~J. Rudner, Yeming Wen, Florian Wenzel, Kevin Murphy, D. Sculley, Balaji
  Lakshminarayanan, Jasper Snoek, Yarin Gal, and Dustin Tran.
\newblock Uncertainty {Baselines}: {Benchmarks} for {Uncertainty} \&
  {Robustness} in {Deep} {Learning}.
\newblock {\em arXiv:2106.04015 [cs]}, June 2021.
\newblock arXiv: 2106.04015.

\bibitem{radford_learning_2021}
Alec Radford, Jong~Wook Kim, Chris Hallacy, Aditya Ramesh, Gabriel Goh,
  Sandhini Agarwal, Girish Sastry, Amanda Askell, Pamela Mishkin, Jack Clark,
  Gretchen Krueger, and Ilya Sutskever.
\newblock Learning {Transferable} {Visual} {Models} {From} {Natural} {Language}
  {Supervision}.
\newblock {\em arXiv:2103.00020 [cs]}, Feb. 2021.
\newblock arXiv: 2103.00020.

\bibitem{ramesh_hierarchical_2022}
Aditya Ramesh, Prafulla Dhariwal, Alex Nichol, Casey Chu, and Mark Chen.
\newblock Hierarchical {Text}-{Conditional} {Image} {Generation} with {CLIP}
  {Latents}, Apr. 2022.
\newblock arXiv:2204.06125 [cs].

\bibitem{dalle2}
Aditya Ramesh, Prafulla Dhariwal, Alex Nichol, Casey Chu, and Mark Chen.
\newblock Hierarchical text-conditional image generation with clip latents,
  2022.

\bibitem{rudin_interpretable_2021}
Cynthia Rudin, Chaofan Chen, Zhi Chen, Haiyang Huang, Lesia Semenova, and Chudi
  Zhong.
\newblock Interpretable {Machine} {Learning}: {Fundamental} {Principles} and 10
  {Grand} {Challenges}.
\newblock {\em arXiv:2103.11251 [cs, stat]}, July 2021.
\newblock arXiv: 2103.11251.

\bibitem{russakovsky_imagenet_2015}
Olga Russakovsky, Jia Deng, Hao Su, Jonathan Krause, Sanjeev Satheesh, Sean Ma,
  Zhiheng Huang, Andrej Karpathy, Aditya Khosla, Michael Bernstein,
  Alexander~C. Berg, and Li Fei-Fei.
\newblock {ImageNet} {Large} {Scale} {Visual} {Recognition} {Challenge}.
\newblock {\em arXiv:1409.0575 [cs]}, Jan. 2015.
\newblock arXiv: 1409.0575.

\bibitem{salman_unadversarial_2021}
Hadi Salman, Andrew Ilyas, Logan Engstrom, Sai Vemprala, A. Madry, and Ashish
  Kapoor.
\newblock Unadversarial {Examples}: {Designing} {Objects} for {Robust}
  {Vision}.
\newblock In {\em {NeurIPS}}, 2021.

\bibitem{prompttext2}
Timo Schick and Hinrich Sch{\"{u}}tze.
\newblock True few-shot learning with prompts - {A} real-world perspective.
\newblock {\em CoRR}, abs/2111.13440, 2021.

\bibitem{shridhar_cliport_2021}
Mohit Shridhar, Lucas Manuelli, and Dieter Fox.
\newblock {CLIPort}: {What} and {Where} {Pathways} for {Robotic}
  {Manipulation}.
\newblock {\em arXiv:2109.12098 [cs]}, Sept. 2021.
\newblock arXiv: 2109.12098.

\bibitem{flava}
Amanpreet Singh, Ronghang Hu, Vedanuj Goswami, Guillaume Couairon, Wojciech
  Galuba, Marcus Rohrbach, and Douwe Kiela.
\newblock {FLAVA:} {A} foundational language and vision alignment model.
\newblock {\em CoRR}, abs/2112.04482, 2021.

\bibitem{florence}
Lu Yuan, Dongdong Chen, Yi{-}Ling Chen, Noel Codella, Xiyang Dai, Jianfeng Gao,
  Houdong Hu, Xuedong Huang, Boxin Li, Chunyuan Li, Ce Liu, Mengchen Liu,
  Zicheng Liu, Yumao Lu, Yu Shi, Lijuan Wang, Jianfeng Wang, Bin Xiao, Zhen
  Xiao, Jianwei Yang, Michael Zeng, Luowei Zhou, and Pengchuan Zhang.
\newblock Florence: {A} new foundation model for computer vision.
\newblock {\em CoRR}, abs/2111.11432, 2021.

\bibitem{zamir_taskonomy_2018}
Amir Zamir, Alexander Sax, William Shen, Leonidas Guibas, Jitendra Malik, and
  Silvio Savarese.
\newblock Taskonomy: {Disentangling} {Task} {Transfer} {Learning}.
\newblock {\em arXiv:1804.08328 [cs]}, Apr. 2018.
\newblock arXiv: 1804.08328.

\end{thebibliography}
}

\clearpage
\appendix
\label{sec:appendix}

% \section{}
\section{Task Bias Probing - Dataset Construction}

    We constructed the datasets for the task bias probing experiments by combining a subset of the OpenImages-V6 dataset, which contains information about objects, attributes and relationships with other objects, \cite{kuznetsova_open_2020,benenson_large-scale_2019} with independently annotated scene text labels associated with a subset of the OpenImages-V5 dataset \cite{krylov_open_2021}. In order to piece together this dataset, we installed the dataset using the recommended FiftyOne \cite{fiftyone} library and combined images from all of the splits in their OpenImages installation. We reported results on four different comparisons: Object v. Scene Text, People v. Actions,  Objects v. Actions and Scene Text v. Actions respectively. Below, we outline the pairwise dataset creation process for each one of these.
    
    \textbf{Objects v. Scene Text}: 
    We consider the intersection of images with scene text labels and images with object detection labels. Images often contain more than one object, but the goal of image classification for object recognition is to identify the most significant one, constraining the task to one label. Similarly to the ImageNet dataset \cite{russakovsky_imagenet_2015}, we want to consider the most salient object in the scene for this label. For every object with detection labels in an image, we calculate the area of the associated bounding box and choose the object label with the maximum area for a given image ID as the label for the object recognition task. (We consider detection labels instead of categorization labels for this reason; categorization labels do not give us any sense of how significant a given label is.) We replace all gendered object labels with the generic human label ``Person''. This leaves us with a set of 175335 images with both object and scene text labels. 
    % Since the set of images with scene text labels is much smaller than the size of OpenImages, we first filter those images for which we have an associated scene text label from the independent scene text dataset. Then we filter this smaller set of images for those which contain object detection labels, retaining all possible object labels. At evaluation time, we want to identify the most significant object inside the scene, so for every object level we also calculate the area of the associated bounding box, keeping only the single object label with the maximum area for a given ID. We replace all gendered object labels with the generic human label 'Person'. This leaves us with a set of images with 175335 images paired with a single scene text and object label. 
    
    \textbf{People v. Actions}: 
    For this comparison, we consider all images with action annotations in OpenImages. (All such images must contain people by definition of the action recognition task, so we do not need to consider an intersection.) As of V6, OpenImages does have human action annotations \cite{kuznetsova_open_2020}; unfortunately, there is no field for `actions' in the dataset. 
    
    Instead, actions are distributed through a couple of different fields. Images that contain actions form a subset of those labeled with `relationships' in OpenImages. Every relationship instance is defined as a 3-tuple of the form (first label, relationship label, second label). The relationship label is a general term for the word that connects the first and second labels and can take various forms. For example, if `is' is the relationship label, the second label can be an attribute of the first label. Alternatively, the relationship label could be an unconjugated verb, in which case the second label is another object. For all the relationships available, we first filter the entire dataset to only include those where the first label refers to a person-related class (In OpenImages, these are  `Boy', `Girl', `Man', `Woman' and `Person') to ensure action recognition is a valid task on the datapoints. We then isolate those pairs where the relationship label directly defines an action (eg. `read', `dance') and those where the relationship label is generic (eg. `is') followed by  a verb-like attribute (eg. `Cry', `Jump'). In the former case, we take the relationship label as the action label and in the latter case, we use the second label as the action. We conjugate all verbs in their present continuous form. The process leaves us with the first label (which is guaranteed to be a person) and an associated action label for a set of images. We then remove duplicates. We sample from each action to correct for label imbalance. Our final dataset contains 89626 distinct images.

    \textbf{Objects v. Actions}: 
    We obtain a set of images containing actions per the previous section. Among these, we consider images that have object detection labels to obtain images where people interact with another object through the action. We further discard any cases where the second label in the relationship 3-tuple is also a person, leaving us with 8611 images with paired action and inanimate object labels. 
    % The goal of this task is to compare the task bias that zero shot vision language models have for actions vs. labels objects that are not people. We follow the same process as in the People v. Actions case \textcolor{blue}{Ishaan says: Is there a better way in which I can phrase this?} to isolate the action valid dataset and get the corresponding action labels. In this case, we also discard all the datapoints of the form where the relationship label is generic (eg. 'is'). This leaves us with a set of points for which people are guaranteed to be in the scene but also interacting with another object through the action, which forms the second label in the relationship instance tuple. We further discard any cases where the second label is also a person, leaving us with 8611 images with paired action and inanimate object labels. 
    
    \textbf{Scene Text v. Actions}: 
    We again consider the subset of images with valid action labels, this time taking the subset of images that also contain scene text. Our final dataset contains 56027 labelled images with both pairs.

    When datasets are used for training prompts and evaluation, $90\%$ of the data is used for training and the remaining $10\%$ is used as a held out set.
    
    \section{Task Bias - Task-Directed Text Prompting Details}
    
    This section provides details for the experiment from Section 3.3. The goal of the experiment is to determine whether a text prompt, added as a prefix to the text choices, can guide the zero-shot classification procedure to solve a desired task. In Table \ref{tab:prompts}, we list the prompts used and the associated intended task. We arrived at these prompts from the original prompts shown to lead to an improvement in baseline performance in CLIP \cite{radford_learning_2021}, reused by ALIGN \cite{jia_scaling_2021} and further models. The primary consideration for designing these prompts was that they must provide sufficient information for a human to understand which task is intended. We experimented with various prompts fulfilling this criterion, choosing the best among them. (This makes it especially surprising that for some experiments, the clarifying additional text information actually results in substantially \textit{worse} performance.)

\begin{table}[]
\centering
\begin{tabular}{@{}ll@{}}
\toprule
\textbf{Task}   & \textbf{Prefix}   \\\midrule
Scene Text      & This is a photo of text which reads                 \\\midrule
Actions         & This is a photo of someone who is     \\\midrule  
Objects, People & This is a photo of \textless{}article\textgreater{} \\\bottomrule
      
\end{tabular}
\caption{Tasks we investigated and the associated prefixes we attach to form our prompts.}
\label{tab:prompts}
\end{table}

    \section{Downstream Task Evaluation - Details}
    
    For the results in Table ~\ref{tab:second_numbers}, we use the model prompted toward the object task on the object vs. scene text dataset. All numbers are reported on our held-out set of this dataset containing ~17000 images. 

    \section{Task Bias -- Task Disambiguation Baselines}
    
    A preliminary question to whether we can resolve task bias is whether we can detect the direction of the task bias in the visual representation solely from the image embedding. In this section, we investigate the effective of using the full attention mask and the input image toward guessing the task bias in an input.
    %Motivated by highly informative visualizations from the self attention maps produced using CLIP's image encoder, we experiment with predicting the task preference labels directly from image and associated artifacts available at encoding time. 
    We provide preliminary baselines for the difficult task of predicting which task a zero-shot model is solving for the Objects v. Scene Text and Actions v. Scene Text pairs. %\textcolor{blue}{Ishaan says: Is it helpful to justify 'why' here?}
    
    \textbf{Dataset}: The results from our task bias probing experiments give us the per label task preference for every image in our dataset. We use these pseudo-labels as indicators of CLIP's task bias on the particular image considered. Therefore, given an image that is part of a pairwise dataset as an input, we repurpose the index of CLIP's preferred task for that image as a label for training the classifier, ensuring that the final test set that we report results on is near-balanced.
    
    \textbf{Methods and Results}: We train four different kinds of classifiers for the two paired datasets. These classifiers differ in their architecture and input space, but in each case the set of labels remains the same. %We explain the classifiers and the baselines below. 
    Our results are summarized in \ref{tab:classifier1} and \ref{tab:classifier2}, reporting the accuracy of the best trained classifier on the test set.
    
    \textit{Frequent} refers to the typical baseline in binary classification that always predicts the label which occurs more frequently inside the test set.
    
    \textit{Image} refers to using the 3-channel input RGB image directly as the input for the classifier. The model used is a ResNet-18 \cite{he_deep_2015} without pre-training.
    
    \textit{Image+Attention Overlay} refers to overlaying the scaled self-attention map from CLIP's image encoder on the image and pre-processing this as a new image. We use the method from \cite{abnar_quantifying_2020,gildenblat_explainability_2021}, termed `attention rollout', to calculate the self attention maps, modifying it for CLIP's architecture. We also experiment with other forms of self-attention maps, including using final layer only, and observe similar results for those. Once we have the attention map, we normalize it with its maximum value, scale it to the image shape, and colorize it using OpenCV's JET colormap \cite{opencv_library} to turn it into an RGB image. Finally, we add this to the RGB image and use it as an input.  The model used is also a ResNet-18 without pre-training.
    
    \textit{Embedding} refers to classifying directly on top of CLIP's 512-dimensional representation of the image. Not that this is a function of the self-attention and the input image, given the structure of transformers. The model used is a shallow 4-layer MLP, with layers sizes [256, 128, 64, 2] respectively.

    \textit{Embedding+Image+Attention} refers to classifying from both the input images and the embedding. Note that we do not necessarily expect this to work better as it gives redundant information to the classifier but in different forms. This is because the embedding itself is a function of the self-attention and the root image. For the model, we use a ResNet18 backbone with a single linear layer to produce a 256 dimensional representation. We further use a single linear layer to constrain the CLIP's image embedding to 256 dimensions. These are then fused and passed through the MLP used for \textit{Embedding} only.
    
    All models are trained end-to-end with the Adam optimizer \cite{kingma_adam_2017}, with learning rate 0.0001 and other hyperparameters set to their default values. 

    The takeaway from these experiments provides some of the motivation for our method. We see that while predicting solely from the image is close to chance, it is relatively much easier to predict the direction of task bias from the embedding.
    
    \begin{table}[]
    \centering
\begin{tabular}{ll}
\toprule
\textbf{Experiment}                & \textbf{Test Accuracy (\%)} \\ \midrule
\textit{Frequent}                  & 54.9                        \\ \midrule
\textit{Image}                     & 61.4                             \\ \midrule
\textit{Image+Attention}           & 62.6                             \\ \midrule
\textit{Embedding}                 & \textbf{71.7}                            \\ \midrule
\textit{Embedding+Image+Attention} & 71.4                             \\ \bottomrule
\end{tabular}
\caption{Results for various classifiers on Objects v. Scene Text task bias clarification task}
\label{tab:classifier1}
\end{table}

\begin{table}[]
\centering
\begin{tabular}{ll}
\toprule
\textbf{Experiment}                & \textbf{Test Accuracy (\%)} \\ \midrule
\textit{Frequent}                  & 53.9                        \\ \midrule
\textit{Image}                     & 59.9                             \\ \midrule
\textit{Image+Attention}           & 61.6                             \\ \midrule
\textit{Embedding}                 & 72.0                             \\ \midrule
\textit{Embedding+Image+Attention} & \textbf{72.8}                            \\ \bottomrule
\end{tabular}
\caption{Results for various classifiers on Actions v. Scene Text task bias clarification task}
\label{tab:classifier2}
\end{table}

\section{Further Attention Examples}

\begin{figure*}[t!]
    \centering
    \includegraphics[width=\linewidth]{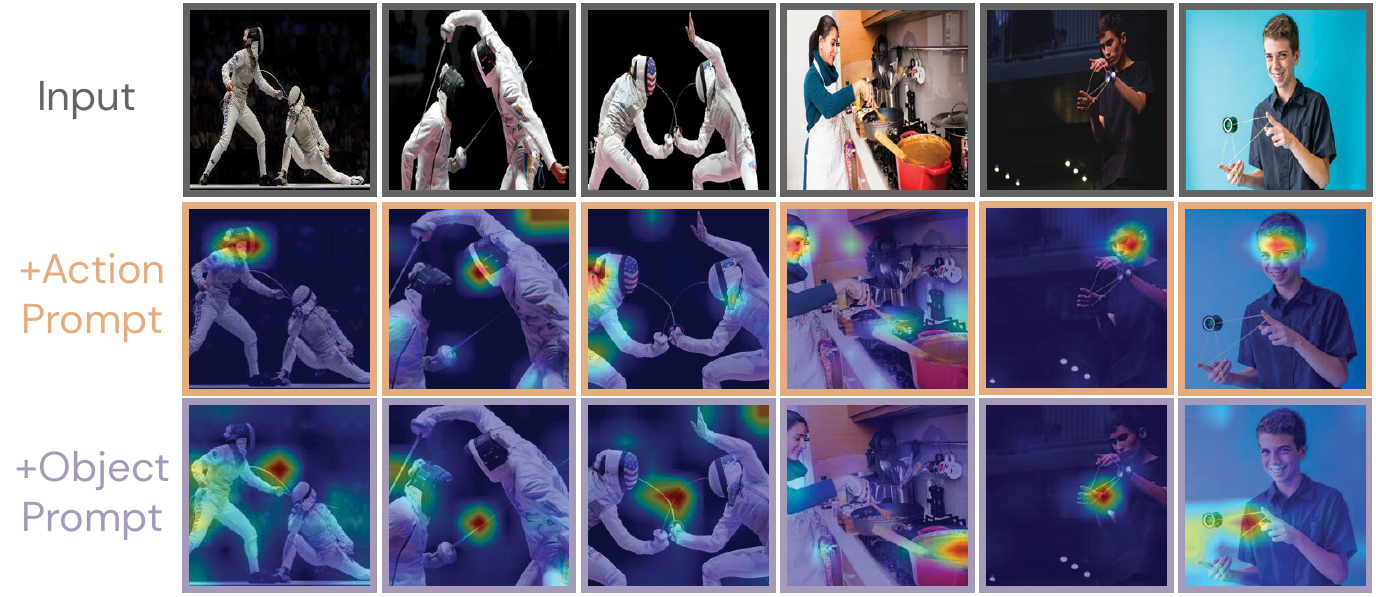}
    \caption{Difference between attention maps with \textcolor{orangeact}{action recognition} and \textcolor{purpobj}{object recognition} prompting on in-the-wild images. Best viewed with zoom. We see the \textcolor{orangeact}{action recognition} prompt results in more attention on people and the \textcolor{purpobj}{object recognition} prompt results in more attention on the objects around them.}
    \label{fig:action_attention_directed}
\end{figure*}

Here we provide some additional examples of the differences between attention maps for prompting in the directions of different tasks, displayed in Fig. \ref{fig:action_attention_directed}. In particular, here we highlight actions, which we did not have room for in the main paper. Note that these examples are all `in-the-wild', not from our dataset. 

% \ifarxiv \clearpage 
% \fi

\end{document}